\documentclass[sigconf]{acmart}
\usepackage{comment}
\def\ie{\emph{i.e.}}

\usepackage{multirow}
\newcommand{\tabincell}[2]{\begin{tabular}{@{}#1@{}}#2\end{tabular}}
\AtBeginDocument{%
  \providecommand\BibTeX{{%
    \normalfont B\kern-0.5em{\scshape i\kern-0.25em b}\kern-0.8em\TeX}}}

\setcopyright{acmcopyright}
\copyrightyear{2023}
\acmYear{2023}
\setcopyright{acmlicensed}
\acmConference[MM '23] {Proceedings of the 31st ACM International Conference on Multimedia}{October 29--November 3, 2023}{Ottawa, ON, Canada.}
\acmBooktitle{Proceedings of the 31st ACM International Conference on Multimedia (MM '23), October 29--November 3, 2023, Ottawa, ON, Canada}
\acmPrice{15.00}
\acmISBN{979-8-4007-0108-5/23/10} 

\acmDOI{10.1145/3581783.3612409}

\definecolor{crncolor}{rgb}{1,.2,0.8}

\definecolor{crnmcolor}{rgb}{0.3,.8,0.2}
\newcommand{\CRNM}[1]{{\textcolor{crnmcolor}{[\textbf{} #1]}}}

\definecolor{cnlcolor}{rgb}{.5,0.1,0.9}

\begin{document}

\title{Bridging Language and Geometric Primitives for Zero-shot Point Cloud Segmentation}

\author{Runnan Chen}
\orcid{1234-5678-9012}
\affiliation{%
  \institution{The University of Hong Kong}
  \country{}
}

\author{Xinge Zhu}
\affiliation{%
  \institution{The Chinese University of Hong Kong}
  \country{}
  }

\author{Nenglun Chen}
\affiliation{%
  \institution{The University of Hong Kong}
  \country{}
}

\author{Wei Li}
\affiliation{%
 \institution{Inceptio}
 \country{}
 }

\author{Yuexin Ma}
\affiliation{%
  \institution{ShanghaiTech University}
  \country{}
  }

\author{Ruigang Yang}
\affiliation{%
  \institution{Inceptio}
  \country{}
  }
\author{Wenping Wang}
\affiliation{%
  \institution{Texas A\&M University}
  \country{}
  }




\begin{abstract}
We investigate transductive zero-shot point cloud semantic segmentation, where the network is trained on seen objects and able to segment unseen objects. The 3D geometric elements are essential cues to imply a novel 3D object type. However, previous methods neglect the fine-grained relationship between the language and the 3D geometric elements. To this end, we propose a novel framework to learn the geometric primitives shared in seen and unseen categories' objects and employ a fine-grained alignment between language and the learned geometric primitives. Therefore, guided by language, the network recognizes the novel objects represented with geometric primitives. Specifically, we formulate a novel point visual representation, the similarity vector of the point's feature to the learnable prototypes, where the prototypes automatically encode geometric primitives via back-propagation. Besides, we propose a novel Unknown-aware InfoNCE Loss to fine-grained align the visual representation with language. Extensive experiments show that our method significantly outperforms other state-of-the-art methods in the harmonic mean-intersection-over-union (hIoU), with the improvement of 17.8\%, 30.4\%, 9.2\% and 7.9\% on S3DIS, ScanNet, SemanticKITTI and nuScenes datasets, respectively. Codes are available\footnote{\url{https://github.com/runnanchen/Zero-Shot-Point-Cloud-Segmentation}.}
\end{abstract}

\keywords{Zero-shot learning, semantic segmentation, point cloud}

\maketitle

\section{Introduction}
Semantic segmentation on the point cloud is a fundamental task in 3D scene understanding, boosting the development of autonomous driving, service robots, digital cities, etc. Although some recent methods \cite{hu2020randla,thomas2019kpconv,cheng20212,xu2021rpvnet,kong2023rethinking,contributors2020mmdetection3d,xu2023human,zhu2021cylindrical,liu2023segment,kong2023robo3d,chen2023studies,kong2023benchmarking} achieve promising performance, they heavily rely on labour-intensive annotations for supervision. By leveraging word embedding as auxiliary information, zero-shot semantic segmentation can recognize the unseen objects whose labels are unavailable. It is beneficial for visual perception in a new scene that contains novel objects. It can also be a pre-annotation tool for automatically labelling the novel objects \cite{michele2021generative,chen2023clip2scene,lu2023see,chen2022towards,chen2023towards,chen2020unsupervised}.

\begin{figure}
  \centerline{\includegraphics[width=0.48\textwidth]{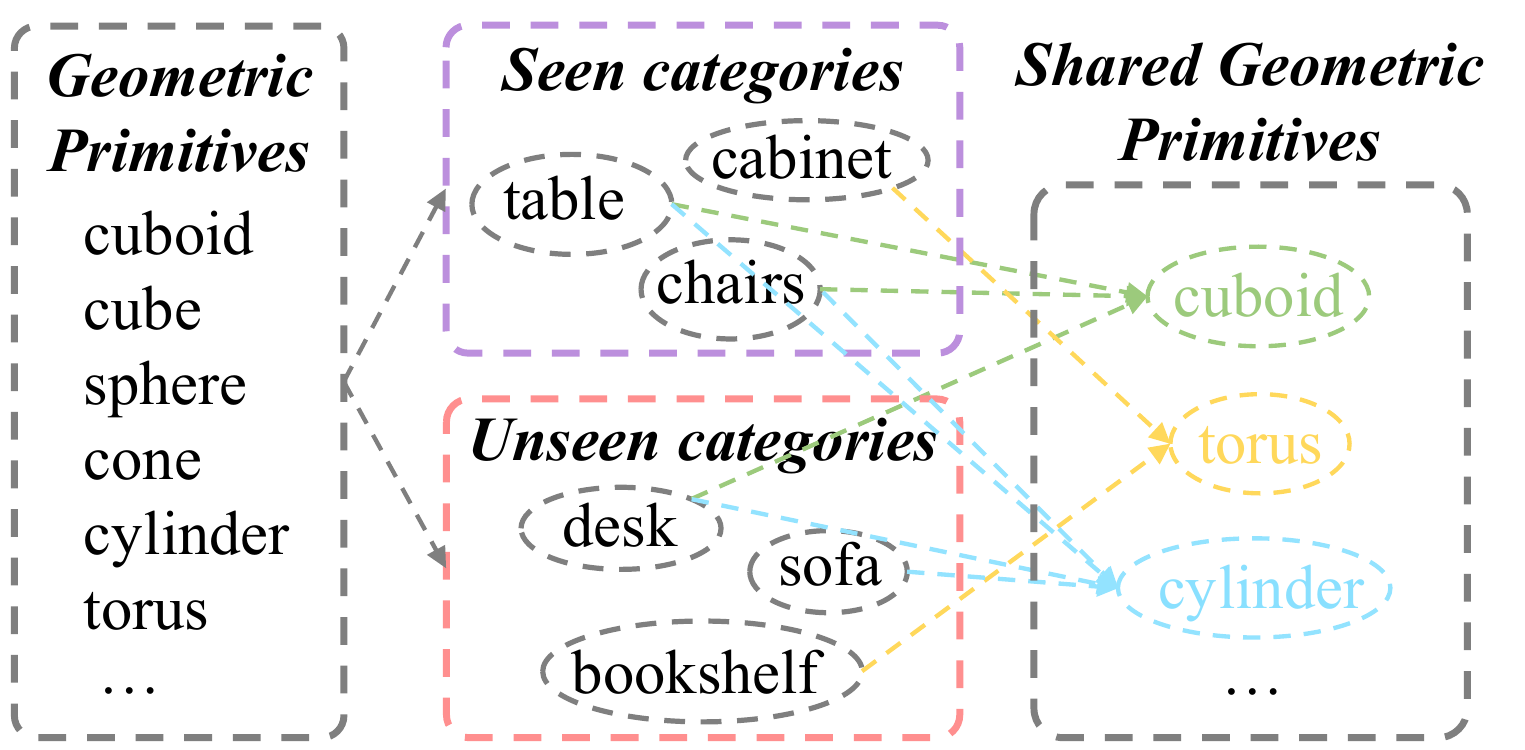}}
  \vspace{-2ex}
  \caption{3D object consists of geometric primitives such as cuboid, cube, cylinder, etc. The 3D geometric elements are essential cues that imply a novel 3D object type. For example, both the table and desk have cuboid and cylinder structures (green and blue dash lines) and similar language semantics (Figure \ref{fig:unknown_aware}). Our method employs fine-grained alignment on language ("table", "desk" et al.) and the geometric primitives shared in seen and unseen categories. Therefore, by the guidance of language, the network is able to recognize the novel object represented with the learned geometric primitives.}
  \label{fig:teaser}
  \vspace{-4ex}
\end{figure}

Zero-shot learning (ZSL) focuses on transferring knowledge from seen to unseen categories. The preliminary ZSL setting predicts only unseen categories, while generalized ZSL (GZSL) predicts both seen and unseen categories. In the aspect of training data, it includes the inductive and transductive settings \cite{zhang2021prototypical,michele2021generative}. Only the seen class samples and labels are available for training the network in the inductive setting. As for the transductive setting, the unlabeled objects of unseen classes are also accessible. In the semantic segmentation scenario, transductive GZSL is a more common setting for zero-shot segmentation (ZSS) because the seen and unseen categories often appear together in a scene. The ZSS problem we studied belongs to the transductive GZSL.

Impressive progress of ZSS has been achieved on the 2D images \cite{bucher2019zero,gu2020context,li2020consistent,hu2020uncertainty,zhang2021prototypical}. They typically generate the fake features of unseen categories for training the classifier or enhance the structure consistency between the visual features and semantic representation. ZSS is not fully explored in the 3D point cloud scenario. Only one method \cite{michele2021generative} investigates this problem to the best of our knowledge. It generates unseen class features with semantic embeddings for training the classifier. However, the fine-grained relationship between the language and the 3D geometric elements in seen and unseen categories, which are important to reason the unseen object types, are not explicitly considered.




In this paper, we investigate transductive zero-shot segmentation (ZSS) on the 3D point cloud, \ie , the visual features of unseen categories are available during training \cite{zhang2021prototypical,michele2021generative}. Our key observation is that the 3D geometric elements are essential cues that imply a novel 3D object type. (Figure \ref{fig:teaser}). For example, chairs and sofas have similar geometric elements such as armrest, backrest and cushion, and they are also close in the semantic embedding space (Figure \ref{fig:unknown_aware}). Based on the observation, we propose a novel framework to learn the geometric primitives shared in seen and unseen categories and fine-grained align the language semantics with the learned geometric primitives. Specifically, inspired by the bag-of-words model \cite{wallach2006topic,fu2019paraphrase}, we formulate a novel point visual representation that encodes geometric primitive information, i.e., the similarity vector of the point's feature to the geometric primitives, where geometric primitives are a group of learnable prototypes updated by back-propagation. For bridging the language and the geometric primitives, we first conduct the language semantic representation as a mixture-distributed embedding, it is because a 3D object composed of multiple geometric primitives. Besides, the network is naturally biased towards the seen classes, leading to significant misclassifications of the unseen classes (Figure \ref{fig:visual}). To this end, we propose an Unknown-aware InfoNCE Loss that fine-grained aligns the visual and semantic representation at the same time alleviating the misclassification issue. Essentially, it pushes the unseen visual representations away from the seen categories' semantic representations (Figure \ref{fig:unknown_aware}), enabling the network to distinguish the seen and unseen objects. In the inferring state, under the guidance of semantic representations, a novel object is represented with learned geometric primitives and can be classified into the correct class.

Extensive experiments conducted on S3DIS, ScanNet, Semantic-KITTI and nuScenes datasets show that our method outperforms other state-of-the-art methods, with the improvement of 17.8\%, 30.4\%, 9.2\% and 7.9\%, respectively.

The contributions of our work are as follows.
\begin{itemize}
\item {To solve the transductive zero-shot segmentation on the 3D point cloud, we propose a novel framework to model the fine-grained relationship between language and geometric primitives that transfer the knowledge from seen to unseen categories.}
\item {We propose an Unknown-aware InfoNCE Loss for the fine-grained visual and semantic representation alignment among seen and unseen categories.}
\item {Our method achieves state-of-the-art performance for zero-shot point cloud segmentation on S3DIS, ScanNet, Semantic-KITTI and nuScenes datasets.}
\end{itemize}

\section{Related Work}

\begin{figure*}
  \centerline{\includegraphics[width=1\textwidth]{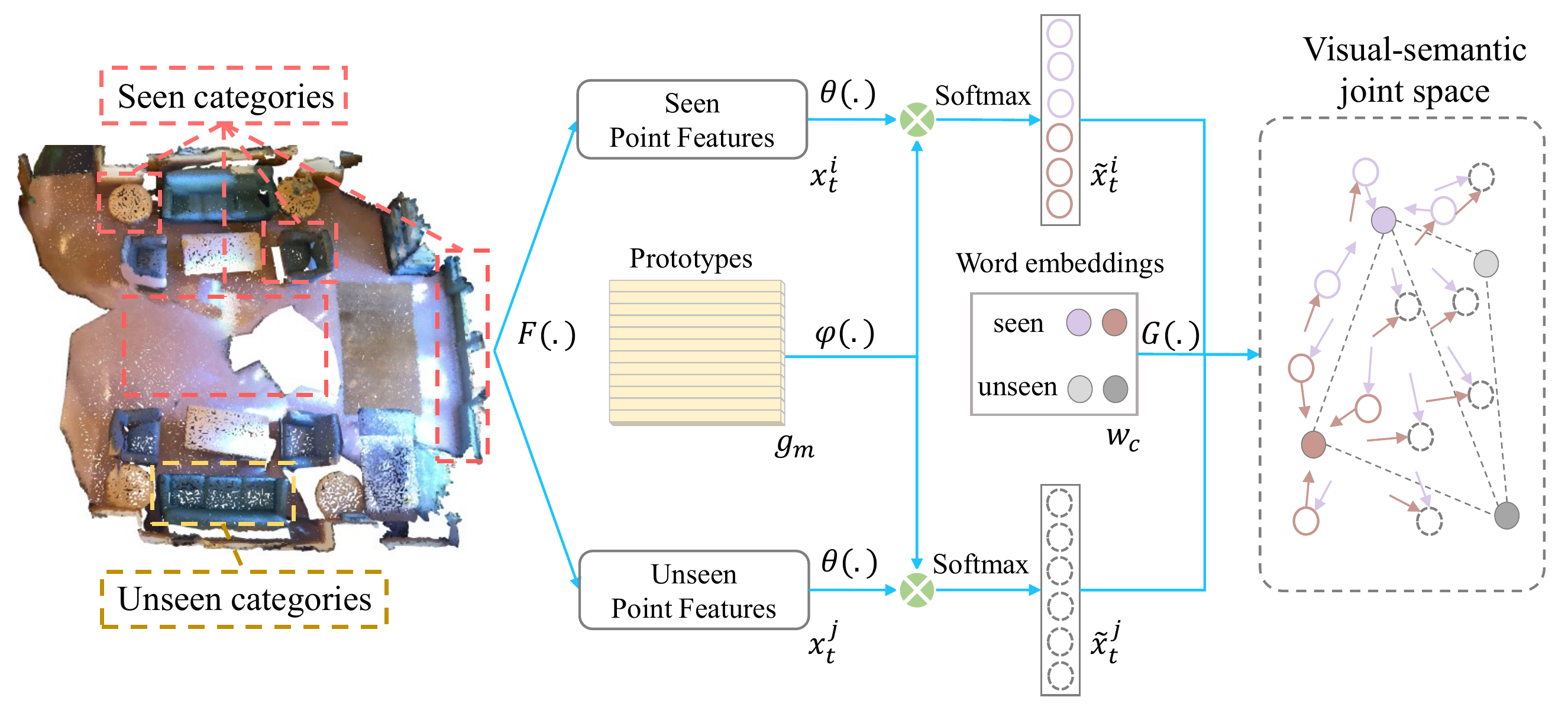}}
  \vspace{-3ex}
  \caption{Illustration of the overall framework. Our framework contains two modules in one end-to-end training process. Firstly, we obtain the point-wise feature for seen categories $\{x_t^i\}_{t=1}^{T_i}$ and unseen categories $\{x_t^j\}_{t=1}^{T_j}$. The point visual representation (hollow circles) is formulated as the similarity vector of its feature to the prototypes $\{g_m\}_{m=1}^{M}$, which is obtained by the cross attention operation between prototypes and point features and is regulated by the softmax operation with inverse temperature. Next, we generate semantic representation (solid circles) from word embedding via $G(\cdot)$, and align the visual and semantic representations of both seen (coloured) and unseen (grey) category points in the visual-semantic joint space (grey dash box).}
  \label{fig:framework}
  \vspace{-1ex}
\end{figure*}

\subsection{Zero-Shot Segmentation on 2D images}
Zero-shot semantic segmentation (ZSS) is dominated by generalized zero-shot learning because the objects of seen and unseen categories often appear in a scene together. ZS3Net \cite{bucher2019zero} generates pixel-wise fake features from semantic information of unseen classes and then integrate the real feature of seen classes for training the classifier. Gu et al. \cite{gu2020context} further improve ZS3Net by introducing a contextual module that generates context-aware visual features from semantic information. Li et al. \cite{li2020consistent} propose a Consistent Structural Relation Learning (CSRL) approach to model category-level semantic relations and to learn a better visual feature generator. However, they improperly use each unseen ground truth pixel locations for fake feature generation. Hu et al. \cite{hu2020uncertainty} promote the performance by alleviating noisy and outlying training samples from seen classes with Bayesian uncertainty estimation. There is an obvious bias between real and fake features that hinder the knowledge transferring from seen classes to unseen classes. Zhang et al. \cite{zhang2021prototypical} replace the unseen objects with other images to generate training samples and perform segmentation with prototypical matching and open set rejection. Lv et al. \cite{lv2020learning} mitigate this issue using a transductive setting that 
uses both labelled seen images and unlabeled unseen images for training. In this paper, we follow the transductive setting that leverages unseen objects' features for supervision, and the ground truth pixel locations of individual unseen objects are not accessible, which naturally meets the semantic segmentation.

\subsection{Zero-shot Learning on 3D Point Cloud}
Unlike promising progress of zero-shot learning on 2D images, there are few studies conducted on the 3D point cloud. Some methods \cite{cheraghian2019mitigating,cheraghian2020transductive,cheraghian2021zero,cheraghian2019zero} are to study point cloud classification. Cheraghian et al. \cite{cheraghian2019zero} adapts pointNet \cite{qi2017pointnet} to extract object representation and GloVe \cite{pennington2014glove} or W2V \cite{mikolov2013distributed} to obtain semantic information for reasoning the unseen object's types. Cheraghian et al. \cite{cheraghian2019mitigating} adapt GZSL setting and propose a loss function composed of a regression term \cite{zhang2017learning} and a skewness term \cite{radovanovic2010hubs,shigeto2015ridge} to alleviate the hubness problem, which indicates that the model may predict only a few target classes for most of the test instances. Cheraghian et al. \cite{cheraghian2020transductive} further improves \cite{cheraghian2019mitigating} by using a triplet loss. To the best of our knowledge, only one method \cite{michele2021generative} is proposed for semantic segmentation that generates fake features with class prototypes for training the classifier. However, they do not explicitly consider the 3D geometric elements shared across the seen and unseen categories, which are important cues to align the 3D visual features and semantic embedding. In this paper, our method learns geometric primitives that transfer the knowledge from seen to unseen categories for more accurate reasoning of unseen category objects. Besides, instead of generating fake features for the unseen classes, we naturally leverage unseen visual features extracted from the backbone network, which is more intuitive and natural under the transductive setting.

\section{Method}
\subsection{Problem Definition}
In zero-shot semantic segmentation, we leverage word embedding as auxiliary information to segment the unseen categories' objects, whose labels are unavailable during training. Suppose $X, W, Y$ indicates the visual features, semantic information (word embedding of the class name) and ground truth label, respectively. The training set is $D_{train}^{in}=\{(X^{s}_{i}, W^{s}_{i},Y^{s}_{i})^{N^{s}}_{i=1}\}$ for the inductive setting of zero-shot segmentation. In terms of the transductive setting $D_{train}^{trans}=\{(X^{s}_{i}, W^{s}_{i},Y^{s}_{i})^{N^{s}}_{i=1},(X^{u}_{j}, W^{u}_{j})^{N^{u}}_{j=1}\}$, where the superscript $s, u$ stand for seen and unseen categories, respectively, and there is no overlap between the seen and unseen classes.  $N^{s},N^{u}$ is the number of samples that contains seen and unseen categories, respectively. Suppose $i$-th sample have $T_{i}$ points, $x_t^i \in X_{i}$ and $y_t^i \in Y_{i}$ indicates $t$-th point's visual feature and ground truth label.

In this paper, we focus on the transductive zero-shot segmentation, which is more common in the semantic segmentation task because the seen and unseen categories often appear together. Notably, ``unseen" here indicates that visual features are accessible while the labels are unavailable. The goal is to generates pixel-level segmentation masks for both seen and unseen categories. 


\subsection{Approach Overview}
Our framework is illustrated in Figure \ref{fig:framework}. The training stage contains two modules, \ie, Visual Representation with Geometric Primitives and Fine-grained Alignment on Visual and Semantic
Representation. Firstly, we extract the point-wise feature of the seen and unseen categories and formulate the point visual representation as the similarity vector between geometric primitives and point features. Next, we perform fine-grained alignment on visual and semantic representation for both seen and unseen categories via an Unknown-aware InfoNCE Loss. In the inferring stage, our method represents unseen category points with the learned geometric primitives and infers the specific unseen class under the guidance of the semantic representations. In what follows, we present these modules in detail.

\subsection{Visual Representation with Geometric Primitives}
\begin{figure}
  \centerline{\includegraphics[width=0.47\textwidth]{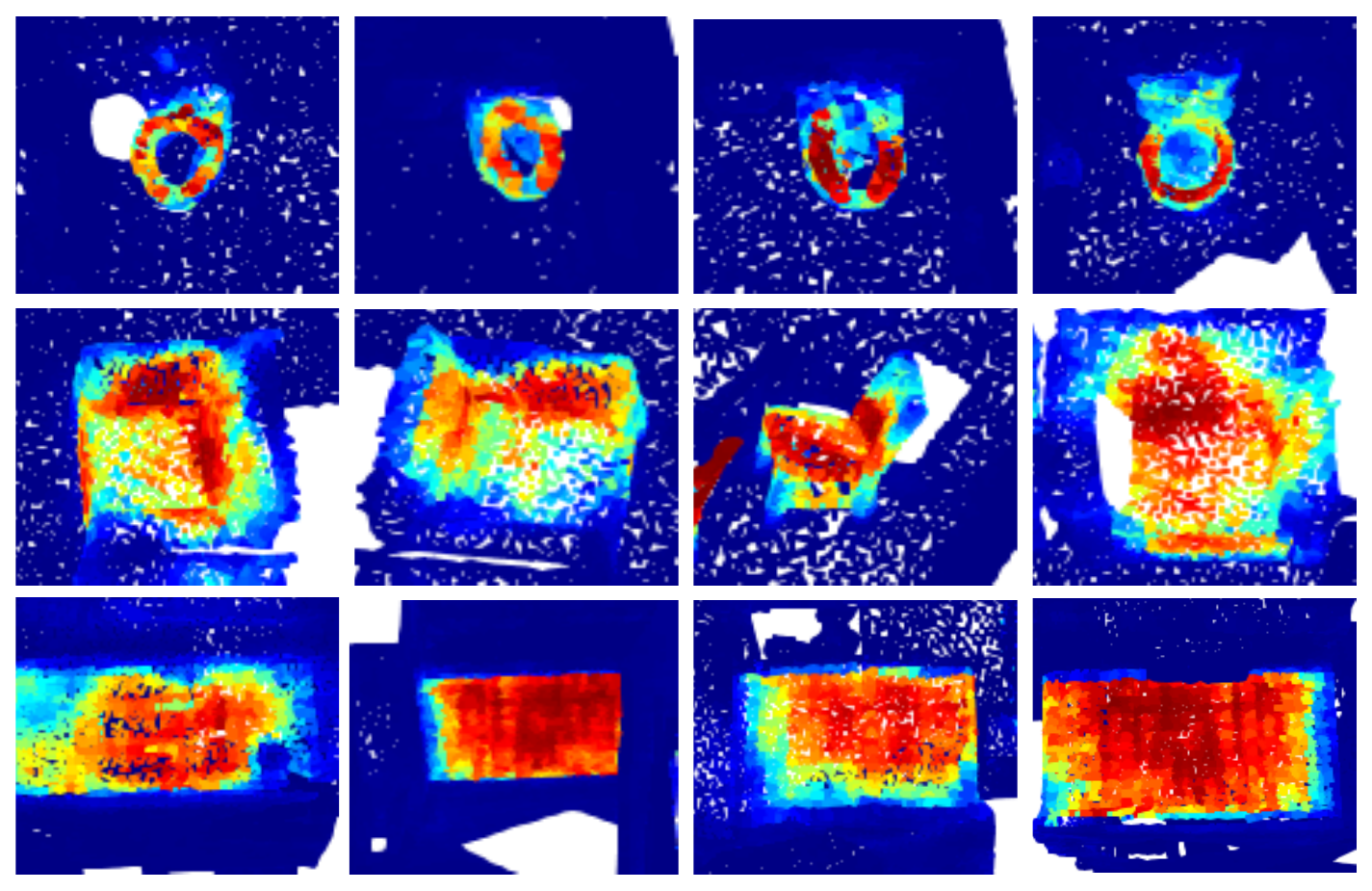}}
  \vspace*{-2ex}
  \caption{Illustration of the learned geometric primitives. From the first to last row are torus, corner, and cuboid, respectively.}
  \label{fig:prototypes}
  \vspace{-2ex}
\end{figure}

Unlike 2D images in the projected view, the 3D object contains complete geometric information. We observe that 3D shapes are composed of common 3D structure elements, such as the cube, cuboid, sphere, cone, cylinder, pyramid, torus, etc. Thus, our basic idea is to disentangle the 3D objects into a set of learnable geometric primitives (Figure \ref{fig:prototypes}). The geometric primitives could constitute a novel 3D object. For example, a simple chair consists of a cushion (cuboid) and four legs (cylinders). Therefore, our method first learns the shared 3D geometric elements in seen and unseen categories and then takes them to represent novel objects.

Inspired by the bag-of-words model \cite{wallach2006topic,fu2019paraphrase} that a sentence can be a dictionary, and its semantic depends on each word's frequency. We formulate point visual representation with geometric primitives, \ie, take geometric primitives as words. A point can be a dictionary where the ``word frequency" is the similarity of the point feature to the corresponding geometric primitive.
\begin{equation}\label{equ:point_representation}
\hat{x}_{t}^i = \{g_{1}:\alpha_{t,1}^i,g_{2}:\alpha_{t,2}^i,g_{3}:\alpha_{t,3}^i,...,g_{M}:\alpha_{t,M}^i\},
\end{equation}
where $\{g_m\}_{m=1}^{M}$ are $M$ learned geometric primitives. $\{\alpha_{t,m}^i\}_{m=1}^{M}$ are the weights of the corresponding $g_{m}$, defined:

\begin{equation}\label{equ:weights}
\begin{split}
\alpha_{t,m}^i &= \frac{\exp(\lambda*d(\theta(x_t^i), \varphi(g_{m})))}{\sum^{M}_{m=1}\exp(\lambda*d(\theta(x_t^i), \varphi(g_{m})))},
\end{split}
\end{equation}
where $x_t^i \in X_{i}$ is the $t$-th point visual feature. $d(\cdot)$ measures the similarity between the point feature $x_t^i$ and the $m$-th geometric primitives $g_{m}$. We utilize dot product operation in this work. $\theta(\cdot)$ and $\varphi(\cdot)$ denote the key and the query function, respectively. $\lambda$ is the inversed temperature term. Note that $\alpha_{t,m}^{i}$ is positive and the sum of possibilities is restricted to 1 to prevent network deterioration, \ie , $\sum^{M}_{m=1}\alpha_{t,m}^{i}=1$.

As the geometric primitives are shared for all categories, we simplify the point representation to be a $M$ dimensional vector:
\begin{equation}\label{equ:final_point_representation}
\tilde{x}_{t}^i = [\alpha_{t,1}^i,\alpha_{t,2}^i,\alpha_{t,3}^i,...,\alpha_{t,M}^i].
\end{equation}
Essentially, $\tilde{x}_{t}^i$ is a possibility distribution that indicates the proportion of the geometric primitives. In this way, a similar distribution implies the same category, and distinct primitives distribution indicates different categories.  

\subsection{Fine-grained Alignment on Visual and Semantic Representation}
In this section, we perform fine-grained alignment on the visual and semantic representation in the aspect of the semantic representation formulation and the Unknown-aware InfoNCE Loss design. In the end, we present how the network infers a scene for both seen and unseen objects.
\paragraph{\textbf{Semantic Representation of Mixture Distribution}}

3D object's visual representation should be the mixture distribution as it consists of multiple geometric primitives. Therefore, we generate the semantic representation $\tilde{w}_{c}$ with multi-kernels from the word embeddings for the fine-grained alignment of the visual-semantic representation.

\begin{equation}\label{equ:semantic_representation}
\begin{aligned}
&\tilde{w}_{c} = \sum_{k=1}^{K}\hat{w}_{c,k}, \hat{w}_{c,k} = G_{k}(w_{c}),
\end{aligned}
\end{equation}

where $\{w_c\}_{c=1}^{C^s+C^u}$ are the word embeddings. They are the output embeddings of the word2vec \cite{mikolov2013efficient} or glove \cite{pennington2014glove} model with the input of categories' names. $C^s$ and $C^u$ are the number of seen and unseen categories, respectively. $K$ is the number of kernels. $\{G_{k}(\cdot)\}_{k=1}^{K}$ are the generation networks.

\begin{figure}
  \centerline{\includegraphics[width=0.45\textwidth]{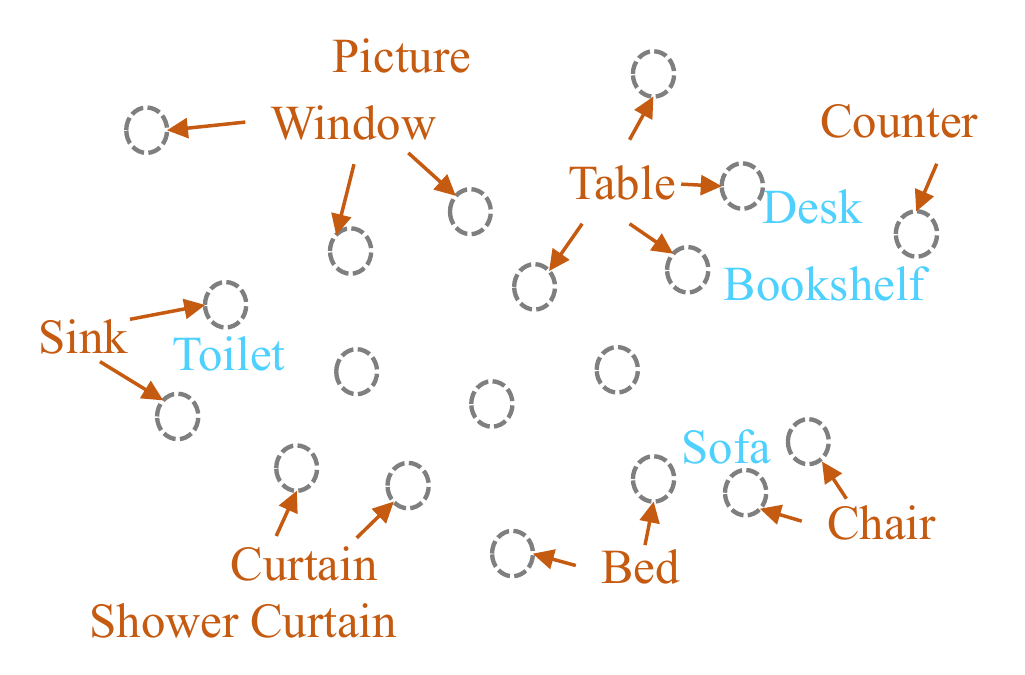}}
  \vspace*{-2.5ex}
  \caption{Illustration of the Unknown-aware InfoNCE Loss for unseen point supervision. The words indicate the W2V+GloVe word embeddings and are represented with t-SNE visualizations \cite{michele2021generative}, where the distance reflects the semantic similarity. The orange and blue words are the seen and unseen categories, respectively. The grey dash circles are the unseen point visual representations. We push the unseen point’s visual representations away from the seen categories’ semantic representations, alleviating the misclassification issue.}
  \label{fig:unknown_aware}
  \vspace{-3ex}
\end{figure}

\begin{table*}[h]
	\centering
	\caption{Comparison with state-of-the-art methods on S3DIS, ScanNet, semanticKITTI and nuScenes dataset by evaluation metrics of mIoU and hIoU. $\dagger$ indicates the adaptation of 2D methods to 3D point clouds, reported in \cite{michele2021generative}. “Superivised$\dagger$” indicates the fully supervised version reported in \cite{michele2021generative}, while “Supervised” is with our backbone network.}\label{tab:overall}
	\scalebox{0.97}{
	\begin{tabular}{l|c c c c | c c c c | c c c c | c c c c}

        \hline
        
		\multirow{2}*{\tabincell{c}{Model}} &
		\multicolumn{4}{c}{S3DIS} & \multicolumn{4}{c}{ScanNet} & \multicolumn{4}{c}{SemanticKITTI} & \multicolumn{4}{c}{NuScenes}\\
        \cline{2-17}
		~ & seen& unseen & all & hIoU & seen& unseen & all & hIoU & seen& unseen & all & hIoU & seen& unseen & all & hIoU\\
        \hline
		Superivised$\dagger$ & 74.0 & 50.0 & 66.6 & 59.6 & 43.3 & 51.9 & 45.1 & 47.2 & 59.4 & 50.3 & 57.5 & 54.5 & 68.1 & 73.2 & 69.4 & 70.5\\
		Superivised & 71.8 & 25.9 & 57.7 & 38.1 & 56.4 & 69.8 & 59.1 & 62.4 & 59.2 & 58.4 & 59.1 & 58.8 & - & - & - & -\\

        \hline
		ZSLPC$\dagger$\cite{cheraghian2019zero}&5.2& 1.3 & 4.0 & 2.1 & 16.4  & 4.2 & 13.9 & 6.7 & 26.4 & 10.2 & 21.8 & 14.7 & 31.7 & 5.3 & 25.1 & 16.3\\
		Devise$\dagger$ \cite{frome2013devise}&3.6& 1.4 & 3.0 & 2.0 & 12.8 & 3.0 & 10.9 & 4.8 & 42.9 &4.2 & 27.6 &7.5 & 51.4 & 4.2 & 41.5 & 7.8\\
		ZS5Net \cite{bucher2019zero} &56.6& 4.8 & 40.6 & 8.8 & 56.8 & 3.9 & 46.2 & 7.3 & 53.2 & 5.1 & 43.1 & 9.3 & 63.5 & 3.2 & 48.4 & 6.1\\
		PMOSR \cite{zhang2021prototypical} &57.4& 6.3 & 41.6 & 11.4 & 55.3 & 5.4 & 45.3 & 9.8 & 55.1 & 8.7 & 45.3 & 15.0 & 64.3 & 3.6 & 51.2 & 6.5\\
		3DGenZ \cite{michele2021generative}&53.1& 7.3 & 39.0 & 12.9 & 32.8 & 7.7 & 27.8 & 12.5 & 41.4 & 10.8 & 35.0 & 17.1 & 67.2 & 3.1 & 51.2 & 5.9\\

		Ours &\textbf{60.4}& \textbf{20.6} & \textbf{48.2} & \textbf{30.7} & \textbf{57.9} & \textbf{34.1} & \textbf{53.1} & \textbf{42.9} & \textbf{54.6} & \textbf{17.3} & \textbf{46.7} & \textbf{26.3} & \textbf{65.7} & \textbf{14.8} & \textbf{53.0} & \textbf{24.2}\\
		\hline

	\end{tabular}}
\end{table*}

\paragraph{\textbf{Unknown-aware InfoNCE Loss}}
We propose an unknown-aware InfoNCE Loss to model the fine-grained relationship between the visual (geometric primitive distribution) and the semantic representation. The loss function is designed for two purposes: 1) distinguishing the specific seen category, 2) enabling the network to identify whether the object is the seen or unseen category. In what follows, we present the details and insights.

To efficiently learn the geometric primitives shared in seen and unseen classes, we cluster these points of seen classes to ensure they are inner-class compact and inter-class distinguishable. \ie , pull in the corresponding visual and semantic representation while pushing away if they are not. The objective function is as follows:
\begin{equation}\label{equ:seen_loss}
\mathcal{L}_{s} = -\log \sum_{i=1}^{N^{s}}\sum_{t=1}^{T_{i}} \frac{\exp(D(\tilde{x}_{t}^{i}, \tilde{w}_{y_{t}^{i}})/\tau)}{\sum_{c=1}^{C^s+C^u}\exp(D(\tilde{x}_{t}^{i}, \tilde{w}_{c})/\tau)},
\end{equation}
where $\tilde{w}_{y_{t}^{i}}$ is the ground truth semantic representation of the $t$-th point in the $i$-th sample. $\tau$ is the inversed temperature term. $C^{s}$ and $C^{u}$ are the number of seen and unseen categories, respectively. $D(\cdot)$ indicates the similarity function between visual and semantic representation, defined as follows:
\begin{equation}\label{equ:distance_function}
D(\tilde{x}_{t}^{i}, \tilde{w}_c)=\sum_{k=1}^{K}d(\tilde{x}_{t}^{i},\hat{w}_{c,k}),
\end{equation}
where $d(\cdot)$ is doc product operation in this paper.

Since the trained model is naturally biased towards the seen classes, leading to significant misclassifications, especially for those unseen semantic representations similar to some seen classes. For example, suppose the table and desk are the seen and unseen classes, respectively. And their semantic representation is similar in word embedding space (Figure \ref{fig:unknown_aware}). Then, if the network is trained only with the table class, it will probably recognize the desk object to be the table (Figure \ref{fig:visual}). Based on this consideration, we push away the unseen categories' visual representations from the seen categories' semantic representations to alleviate the misclassification issue. The objective function is as follows.
\begin{equation}\label{equ:unseen_loss}
\mathcal{L}_{u} = \sum_{j=1}^{N^{u}}\sum_{t=1}^{T_{j}} \frac{\sum_{c=1}^{C^s}\exp(D(\tilde{x}_{t}^{j}, \tilde{w}_{c}))}{\sum_{\hat{c}=1}^{C^s+C^u}\exp(D(\tilde{x}_{t}^{j}, \tilde{w}_{\hat{c}}))}
\end{equation}
Note that the ground truth label $\{\tilde{w}_{y_{t}^{j}}\}_{j=1,t=1}^{N^u,T_j}$ of unseen points are not accessed. Essentially, $\mathcal{L}_{u}$ contains a softmax operation on the possibility distribution among seen and unseen classes. It enforces the possible sum of seen classes to be 0 and the possible sum of unseen classes to be 1. The final loss function is the combination of two items:
\begin{equation}\label{equ:final_loss}
\mathcal{L}_{f} = \mathcal{L}_{s} + \mathcal{L}_{u},
\end{equation}

\paragraph{\textbf{Inference}}
When inferring an new scene scan with the point features $\{x_t^l\}_{i=1}^{T_l}$ obtained by the backbone network $F(\cdot)$, where $T_l$ is the number of scene points. The $t$-th point feature $x_t^l$ is transferred to the visual representation $\tilde{x}_t^l$ with formula \ref{equ:point_representation}, \ref{equ:weights}, \ref{equ:final_point_representation}, and its class $C^{\ast}_i$ is determined by the following formula.

\begin{equation}\label{equ:inferring}
C^{\ast}_i = \mathop{\arg\max}_{c}\frac{\exp(D(\tilde{x}_{t}^{l}, \tilde{w}_{c}))}{\sum_{\hat{c}=1}^{C^s+C^u}\exp(D(\tilde{x}_{t}^{l}, \tilde{w}_{\hat{c}}))},
\end{equation}
where $\{\tilde{w}_{\hat{c}}\}_{\hat{c}=1}^{C^s+C^u}$ is the semantic representation for seen and unseen classes.

\section{Experiments}
\subsection{Dataset}
Our method is evaluated on four datasets, including two indoor datasets ScanNet \cite{dai2017scannet} and S3DIS \cite{armeni2017joint}, and two outdoor dataset semanticKITTI \cite{behley2019iccv} and nuScenes \cite{nuscenes2019}.

\paragraph{\textbf{S3DIS}} S3DIS includes 271 scanned rooms with points labelled among 13 classes. We utilize Area 5 as the validation set and use the other five areas as the training set. To comprehensively evaluate the performance, we also conduct zero-shot setups with the different number of unseen classes, constructing the 2-, 4-, 6-class unseen sets. The detailed splits are: 2-sofa/beam, 4-column/window, 6-bookshelf/board. Note that the classes in the unseen set increase incrementally for different setups, e.g., the 4-unseen set contains the 2-unseen set, and the 6-unseen set contains the 4-unseen set.

\paragraph{\textbf{ScanNet}} ScanNet is an indoor scene dataset that contains 1513 scans with annotations for 20 classes, where 1201 scans for training and 312 scans for validation. We conduct zero-shot setups with the different number of unseen classes, including the 2-sofa/desk, 4-bookshelf/toilet, 6-bathtub/bed, 8-curtain/window, 10-door/counter. The classes in the unseen set increase incrementally for different setups.

\paragraph{\textbf{SemanticKITTI}} SemanticKITTI is a large-scale driving-scene dataset for point cloud segmentation. SemanticKITTI contains 22 sequences, where ten sequences for training and sequence 08 for validation. We conduct zero-shot setups with the different number of unseen classes, including the 2-Motorcycle/truck, 4-bicyclist/traffic-sign, 6-car/terrain, 8-vegetation/sidewalk. The classes in the unseen set increase incrementally for different setups.

\begin{table}[h]
	\centering
	\caption{Results on S3DIS dataset with the evaluation metrics of mIoU and hIoU. We show the performance of the setups with the different numbers (2-sofa/beam, 4-column/window, 6-bookshelf/board) of unseen classes. 'Supervised' is trained on labelled seen and unseen categories with our backbone network.}\label{tab:S3DIS}
	\scalebox{1}{
	\begin{tabular}{l|l|c c c c}

        \hline
        \multirow{2}*{\tabincell{c}{setting}} &
		\multirow{2}*{\tabincell{c}{Model}} &
		Seen &
		Unseen & \multicolumn{2}{c}{Overall}\\
        \cline{5-6}
		~&~ & mIoU & mIoU & mIoU & hIoU \\
		\hline
        0&Supervised & - & - & 57.7 & - \\
        \hline
		\multirow{3}*{\tabincell{c}{2}}& Devise$\ddagger$\cite{frome2013devise} & 55.1 & 10.5 & 48.2 & 17.6\\
		~&3DGenZ \cite{michele2021generative} & 50.4 & 9.1 & 44.0 & 15.4\\
		~&Ours & \textbf{60.2} & \textbf{26.3} & \textbf{55.0} & \textbf{36.6}\\
		\hline
		\multirow{3}*{\tabincell{c}{4}}& Devise$\ddagger$\cite{frome2013devise} & 59.0 & 8.7 & 43.5 & 15.2\\
		~&3DGenZ \cite{michele2021generative} & 53.1 & 7.3 & 39.0 & 12.9\\
		~&Ours & \textbf{61.1}& \textbf{22.1} & \textbf{49.1} & \textbf{32.5}\\
		\hline
		\multirow{3}*{\tabincell{c}{6}}& Devise$\ddagger$\cite{frome2013devise} & 68.3 & 3.8 & 38.5 & 7.2\\
		~&3DGenZ \cite{michele2021generative} & 56.4 & 3.5 & 32.0 & 6.6\\
		~&Ours & \textbf{69.5} & \textbf{8.8} & \textbf{41.5} & \textbf{15.6}\\
		\hline

	\end{tabular}}
\end{table}

\begin{table}[h]
	\centering
	\caption{Results on ScanNet dataset with the evaluation metrics of mIoU and hIoU. We show the performance of the setups with the different numbers (2-sofa/desk, 4-bookshelf/toilet, 6-bathtub/bed, 8-curtain/window, 10-door/counter) of unseen classes. 'Supervised' is trained on labelled seen and unseen categories with our backbone network.}\label{tab:ScanNet}
	\scalebox{1}{
	\begin{tabular}{l|l|c c c c}

        \hline
        \multirow{2}*{\tabincell{c}{setting}} &
		\multirow{2}*{\tabincell{c}{Model}} &
		Seen &
		Unseen & \multicolumn{2}{c}{Overall}\\
        \cline{5-6}
		~&~ & mIoU & mIoU & mIoU & hIoU \\
		\hline
        0&Supervised & - & - & 59.1 & - \\
        \hline
		\multirow{3}*{\tabincell{c}{2}}& Devise$\ddagger$\cite{frome2013devise} & 52.1 & 4.0 & 47.3 & 7.4 \\
		~&3DGenZ \cite{michele2021generative} & 33.4 & 12.8 & 31.3 & 18.5\\
		~&Ours & \textbf{58.6} & \textbf{51.6} & \textbf{57.9} & \textbf{54.9}\\
		\hline
		\multirow{3}*{\tabincell{c}{4}}& Devise$\ddagger$\cite{frome2013devise} & 50.9 & 1.9 & 41.1 & 3.6\\
		~&3DGenZ \cite{michele2021generative} & 32.8 & 7.7 & 27.8 & 12.5\\
		~&Ours & \textbf{57.9} & \textbf{34.1} & \textbf{53.1} & \textbf{42.9} \\
		\hline
		\multirow{3}*{\tabincell{c}{6}}& Devise$\ddagger$\cite{frome2013devise} & 48.2 & 2.2 & 34.4 & 4.2\\
		~&3DGenZ \cite{michele2021generative} & 31.2 & 4.8 & 23.3 & 8.3\\
		~&Ours & \textbf{55.2} & \textbf{15.4} & \textbf{43.2} & \textbf{24.1}\\
		\hline
		\multirow{3}*{\tabincell{c}{8}}& Devise$\ddagger$\cite{frome2013devise} &54.9 & 2.0 & 33.7 & 3.9\\
		~&3DGenZ \cite{michele2021generative} & 29.7 & 2.1 & 18.7 & 3.9\\
		~&Ours & \textbf{51.6} & \textbf{12.1} & \textbf{35.8} & \textbf{19.6}\\
		\hline
		\multirow{3}*{\tabincell{c}{10}} & Devise$\ddagger$\cite{frome2013devise} & 51.5 & 0.1 & 25.8 & 0.2\\
		~&3DGenZ \cite{michele2021generative} & 30.1 & 1.4 & 15.8 & 2.7\\
		~&Ours & \textbf{52.5} & \textbf{9.5} & \textbf{31.0} & \textbf{16.1}\\
		\hline
		
	\end{tabular}}
	\vspace{-1.5ex}
\end{table}

\begin{table}[h]
	\centering
	\caption{Results on semanticKITTI dataset. We show the performance of the setups with the different numbers (2-Motorcycle/truck, 4-bicyclist/traffic-sign, 6-car/terrain, 8-vegetation/sidewalk) of unseen classes. 'Supervised' is trained on labelled seen and unseen categories with our backbone network.}\label{tab:semanticKITTI}
	\scalebox{1}{
	\begin{tabular}{l|l|c c c c}

        \hline
        \multirow{2}*{\tabincell{c}{setting}} &
		\multirow{2}*{\tabincell{c}{Model}} &
		Seen &
		Unseen & \multicolumn{2}{c}{Overall}\\
        \cline{5-6}
		~&~ & mIoU & mIoU & mIoU & hIoU \\
		\hline
        0&Supervised & - & - & 59.1 & - \\
        \hline
		\multirow{3}*{\tabincell{c}{2}}& Devise$\ddagger$\cite{frome2013devise} & 53.7 & 7.6 & 48.8 & 13.3\\
		~&3DGenZ \cite{michele2021generative} & 40.9 & 12.4 & 37.9 & 19.0 \\
		~&Ours & 58.3 & 28.8 & 55.2 & 38.6\\
		\hline
		\multirow{3}*{\tabincell{c}{4}}& Devise$\ddagger$\cite{frome2013devise} & 51.4 & 4.2 & 41.5 & 7.8\\
		~&3DGenZ \cite{michele2021generative} & 41.4 & 10.8 & 35.0 & 17.1\\
		~&Ours  & 55.9 & 22.8 & 48.9 & 32.4\\
		\hline
		\multirow{3}*{\tabincell{c}{6}}& Devise$\ddagger$\cite{frome2013devise} & 48.8 & 2.4 & 34.1 & 4.6\\
		~&3DGenZ \cite{michele2021generative} & 40.3 & 6.5 & 29.6 & 11.2\\
		~&Ours  & 53.6 & 13.3 & 40.9 & 21.3\\
		\hline
		\multirow{3}*{\tabincell{c}{8}}& Devise$\ddagger$\cite{frome2013devise} & 49.2 & 0.5 & 28.7 & 1.0\\
		~&3DGenZ \cite{michele2021generative} & 38.3 & 1.3 & 22.7 & 2.5\\
		~&Ours  & 53.2 & 8.6 & 34.4 & 14.8\\
		\hline
		
	\end{tabular}}
\end{table}

\begin{table}[h]
	\centering
	\caption{Results on nuScenes dataset. We show the performance of the setups with the different numbers (2-Motorcycle/trailer, 4-terrain/traffic-cone, 6-bicycle/car, 8-vegetation/sidewalk) of unseen classes. 'Supervised' is trained on labelled seen and unseen categories with our backbone network.}\label{tab:nuScenes}
	\scalebox{1}{
	\begin{tabular}{l|l|c c c c}

        \hline
        \multirow{2}*{\tabincell{c}{setting}} &
		\multirow{2}*{\tabincell{c}{Model}} &
		Seen &
		Unseen & \multicolumn{2}{c}{Overall}\\
        \cline{5-6}
		~&~ & mIoU & mIoU & mIoU & hIoU \\
		\hline
        0&Supervised & - & - & 69.1 & - \\
        \hline
		\multirow{3}*{\tabincell{c}{2}}& Devise$\ddagger$\cite{frome2013devise} & 53.7 & 7.6 & 48.8 & 13.3\\
		~&3DGenZ \cite{michele2021generative} & 67.8 & 4.2 & 59.9 & 7.9 \\
		~&Ours & 58.9 & 26.9 & 54.9 & 36.9\\
		\hline
		\multirow{3}*{\tabincell{c}{4}}& Devise$\ddagger$\cite{frome2013devise} & 51.4 & 4.2 & 41.5 & 7.8\\
		~&3DGenZ \cite{michele2021generative} & 67.2 & 3.1 & 51.2 & 5.9\\
		~&Ours  & 65.7 & 14.8 & 53.0 & 24.2\\
		\hline
		\multirow{3}*{\tabincell{c}{6}}& Devise$\ddagger$\cite{frome2013devise} & 48.8 & 2.4 & 34.1 & 4.6\\
		~&3DGenZ \cite{michele2021generative} & 53.8 & 3.2 & 34.8 & 6.0\\
		~&Ours  & 68.8 & 14.1 & 48.3 & 23.4\\
		\hline
		\multirow{3}*{\tabincell{c}{8}}& Devise$\ddagger$\cite{frome2013devise} & 49.2 & 0.5 & 28.7 & 1.0\\
		~&3DGenZ \cite{michele2021generative} & 36.5 & 2.1 & 19.3 & 4.0\\
		~&Ours  & 68.4 & 13.7 & 41.1 & 22.8\\
		\hline
		
	\end{tabular}}
	\vspace{-1.5ex}
\end{table}

\paragraph{\textbf{NuScenes}} nuScenes contains 16 classes for LiDAR semantic segmentation, where 850 scenes for training and validation, and the remaining 150 scenes are for testing.  We conduct several zero-shot settings with different numbers of unseen classes, including the 2-Motorcycle/trailer, 4-terrain/traffic-cone, 6-bicycle/car, 8-vegetation/sidewalk.

\subsection{Evaluation Metrics}
We report the mean-intersection-over-union (mIoU) in experiments. Besides, following \cite{bucher2019zero,xian2018zero,zhang2021prototypical}, we adopt the harmonic mean of seen mIoU and unseen mIoU (hIoU) to demonstrate the overall performance of ZSS.
\begin{equation}\label{equ:hIoU}
hIoU=\frac{2\times mIoU_{seen} \times mIoU_{unseen}}{mIoU_{seen} + mIoU_{unseen}},
\end{equation}
where $mIoU_{seen}$ and $mIoU_{unseen}$ represents the mean IoU of seen classes and unseen classes respectively.

\subsection{Implementation Details}
We adopt MinkowskiNet14 \cite{choy20194d} as backbone for ScanNet and S3DIS dataset. The point visual feature dimension is set to be 96. The input and output of the key $\theta(\cdot)$ and the query $\varphi(\cdot)$ function are 96 and 16-dimensional vectors, respectively. The generation network $G(\cdot)$ is a two-layer MLP network with the dimension of 96 and $M\times K$, where $M$ and $K$ are the numbers of prototypes and semantic representation kernel, and set to be 128 and 16, respectively. The voxel size of all experiments is set to be 5 cm for efficient training. And the temperature term $\lambda$ is set to be 4. We adopt the codebase Cylinder3D \cite{zhu2021cylindrical} and apply cartesian coordinates for the semanticKITTI dataset. Following \cite{michele2021generative}, we use the W2V+GloVe word embeddings (600-dimensional vector) as the auxiliary semantic information. Our method is built on the Pytorch platform, optimized by Adam with the default configuration. The batch size for the S3DIS, ScanNet, semanticKITTI and nuScenes are 8, 8, 4 and 8, respectively. Training 300 epochs cost 2 hours on four RTX 2080 TI GPUs for the S3DIS dataset, 8 hours for ScanNet and training 20 epochs cost 40 hours for the semanticKITTI dataset, 15 hours for the nuScenes dataset.

\begin{figure*}[h]
  \centerline{\includegraphics[width=1\textwidth]{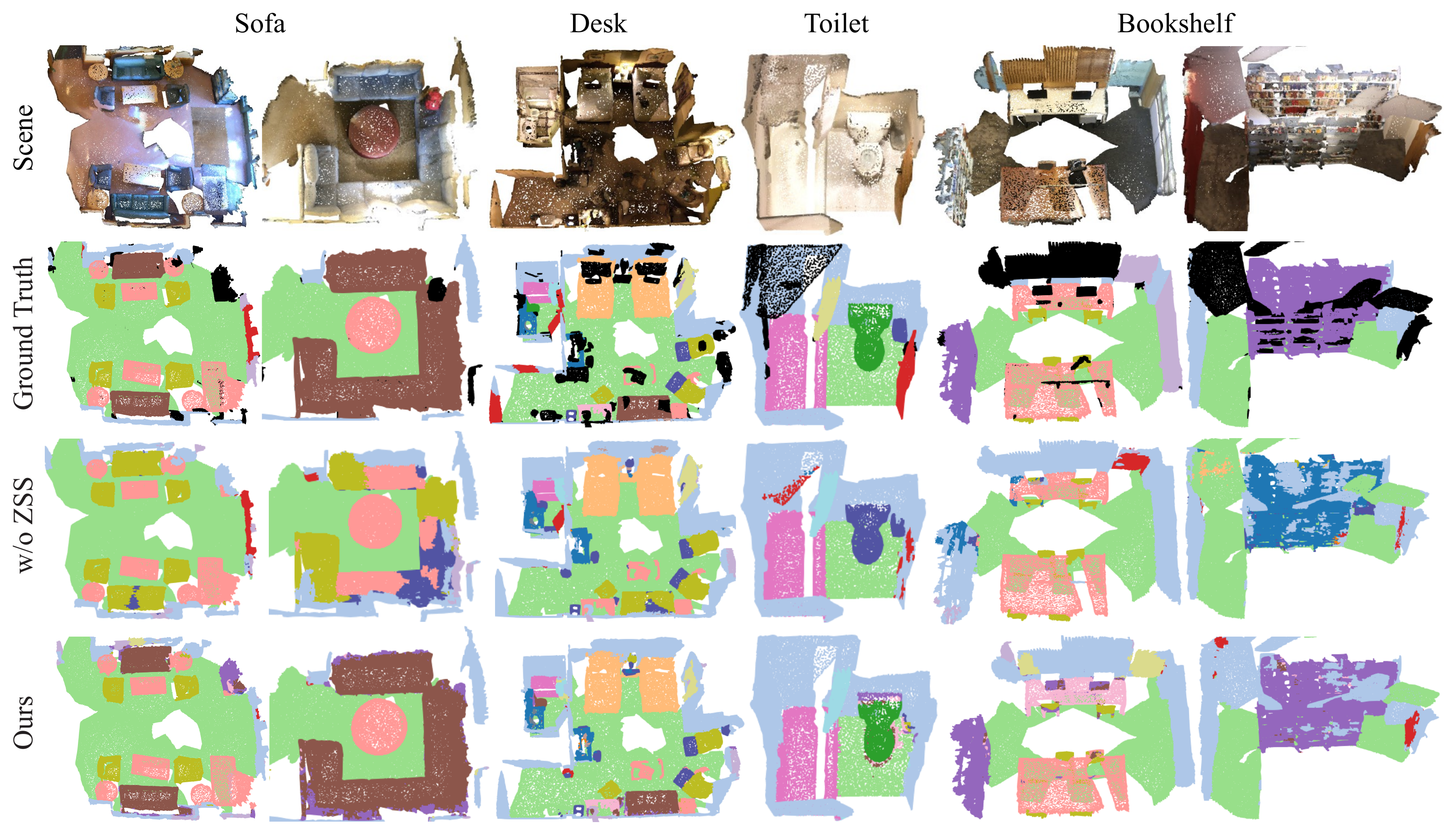}}
  \caption{Qualitative results on ScanNet. From the first row to the last row are input scene (point cloud), ground truth, prediction without ZSS, and our prediction, respectively. From the left column to the right are the unseen classes sofa (brown colour on the ground truth), desk (pink) and toilet (green) and bookshelf (purple). Note that black is the unlabeled region. The model without zero-shot segmentation misclassifies the unseen classes to some seen classes. For example, in the second column, the sofa (the brown region) is misclassified into the table and chair. In comparison, our method achieves decent performance.}
  \label{fig:visual}
\end{figure*}

\subsection{Results and Discussion}
In this section, we report the comparisons with other state-of-the-art methods on four datasets, \ie , S3DIS, ScanNet, SemanticKITTI and NuScenes. Following \cite{michele2021generative}, we benchmark against previous methods in Table \ref{tab:overall}. Besides, follow the zero-shot segmentation on 2D images \cite{bucher2019zero,zhang2021prototypical}, we conduct zero-shot setups with the different number of unseen classes to comprehensively evaluate the performance (Table \ref{tab:S3DIS}, \ref{tab:ScanNet}, \ref{tab:semanticKITTI}, \ref{tab:nuScenes}). We also qualitatively evaluate our method and the counterpart trained only on seen classes (Figure \ref{fig:visual}). In the end, we discuss the limitations and potential directions for future works.

\paragraph{\textbf{Baselines}}
To our knowledge, only one method (3DGenZ \cite{michele2021generative}) investigates the zero-shot point cloud segmentation. We directly compare the method on all settings. Besides, we also adapt three 2D methods \cite{frome2013devise,bucher2019zero,zhang2021prototypical} to 3D point cloud in the transductive setting for a fair comparison, \ie, the backbone is the same as ours, and we pseudo-labelling the unseen points via the Argmax operation on unseen categories and retrain the network with the pseudo labels. Specifically, we directly show the performance of Devise$\dagger$ in \cite{michele2021generative}. We adopt the ZS5Net version in \cite{bucher2019zero}, i.e., the model's own predictions on unseen data is additional pseudo-labelled training data. And the network is trained with labelled seen data and pseudo-labelled unseen data. As for \cite{zhang2021prototypical}, we use the unseen visual features to train the ``unknown visual prototype", named PMOSR in Table \ref{tab:ScanNet}.


\paragraph{\textbf{Comparison with state-of-the-art methods}}
Following \cite{michele2021generative}, we take four classes as unseen categories for three datasets. Specifically, Beam, column, window and sofa are unseen categories for S3DIS datasets; Desk, bookshelf, sofa, toilet for ScanNet datasets; Motorcycle, truck, bicyclist and traffic-sign for SemanticKITTI dataset; and motorcycle, trailer, terrain, traffic-cone for NuScenes dataset. As shown in Table \ref{tab:overall},  our method outperforms other state-of-the-art methods by a large margin, with the hIoU improvement of 17.8\%, 30.4\%, 9.2\% and 7.9\% in four datasets, respectively. Note that our method also works well on the outdoor lidar point cloud dataset (semanticKITTI and nuScenes), showing the generalization ability when the point cloud is sparse.

\begin{table}[h]
	\centering
	\caption{Ablation experiments on ScanNet dataset with 4- unseen class. Base+$\mathcal{L}_{u}$+GP$_{128}$+MK$_{16}$ is our full method.}\label{tab:ablation}
	\scalebox{1}{
	\begin{tabular}{l|c c c c}

        \hline
		\multirow{2}*{\tabincell{c}{Model}} &
		Seen &
		Unseen & \multicolumn{2}{c}{Overall}\\
        \cline{4-5}
		~ & mIoU & mIoU & mIoU & hIoU \\
		\hline
        Base & 54.8 & 0 & 43.8 & 0 \\
        \hline
		Base+$\mathcal{L}_{pseudo}$ & 55.3 & 3.6 & 45.0 & 6.8\\
		Base+$\mathcal{L}_{self}$ & 54.2 & 16.1 & 46.6 & 24.8\\
		Base+$\mathcal{L}_{u}$ & 55.2 & 22.1 & 48.6 & 31.6\\
		\hline
		Base+$\mathcal{L}_{u}$+MK$_{16}$ & 57.3 & 29.4 & 51.7 & 38.9 \\
		\hline
		Base+$\mathcal{L}_{u}$+GP$_{128}$+MK$_{2}$ & 57.1 & 31.6 & 52.0 & 40.7\\
		Base+$\mathcal{L}_{u}$+GP$_{128}$+MK$_{4}$ & 57.9 & 33.1 & 52.9 & 42.1\\
		Base+$\mathcal{L}_{u}$+GP$_{128}$+MK$_{8}$ & 57.7 & 33.3 & 52.8 & 42.2\\
		Base+$\mathcal{L}_{u}$+GP$_{128}$+MK$_{16}$ & 57.9 & 34.1 & 53.1 & 42.9\\
		Base+$\mathcal{L}_{u}$+GP$_{128}$+MK$_{32}$ & 57.1 & 33.1 & 52.3 & 41.9\\
		\hline
		Base+$\mathcal{L}_{u}$+GP$_{48}$+MK$_{16}$ & 53.4 & 33.1 & 49.3 & 40.9\\
		Base+$\mathcal{L}_{u}$+GP$_{96}$+MK$_{16}$ & 54.9 & 32.9 & 50.5 & 41.1\\
		Base+$\mathcal{L}_{u}$+GP$_{128}$+MK$_{16}$ & \textbf{57.9} & \textbf{34.1} & \textbf{53.1} & \textbf{42.9}\\
		Base+$\mathcal{L}_{u}$+GP$_{256}$+MK$_{16}$ & 57.4 & 26.2 & 51.2 & 36.0\\
		\hline
	\end{tabular}}
\end{table}

\paragraph{\textbf{Performance under different number of unseen classes}}
To evaluate the performance, we conduct zero-shot setups with the different numbers of unseen classes in the four datasets. As shown in Table \ref{tab:S3DIS}, \ref{tab:semanticKITTI}, \ref{tab:ScanNet}, \ref{tab:nuScenes}, our method significantly outperforms other state-of-the-art methods, showing the method could handle the different number of unseen classes. As shown in Table \ref{tab:S3DIS}, our method achieves the harmonic mean of seen mIoU and unseen mIoU (hIoU) of 36.6\%, 32.5\% and 15.6\% on 2-, 4- and 6- unseen class, respectively. In the ScanNet 2-, 4-, 6-, 8-, and 10- unseen classes setting, the improved hIoU range from 14\% $\sim$ 36\% (Table \ref{tab:ScanNet}). For the semanticKITTI dataset, our method outperforms other state-of-the-art methods by a large margin, with the mIoU improvement of 19\%, 15\%, 10\% and 12\% in the 2-, 4-, 6- and 8- unseen classes setting (Table \ref{tab:semanticKITTI}). As for NuScenes dataset, the improvemed hIoU is 19\%, 10\%, 12\% and 13\% in the 2-, 4-, 6- and 8- unseen classes setting (Table \ref{tab:nuScenes}). Moreover, we show the upper bound performance with full supervision (refer to setting 0).

Surprisingly, the method achieves high unseen mIoU in the sofa and desk categories (51.6\% mIoU), close to the supervision with ground truth (Table \ref{tab:ScanNet}). The main reason is that sofa and desk have similar semantic representations to the seen classes chair and table, respectively (Figure \ref{fig:unknown_aware}). Constrained by the Unknown-aware InfoNCE Loss, the network concludes that those unknown objects semantically closed to the chair are probably a sofa, and those close to the table are probably a desk. We also show the performance by training on the annotated seen and unseen categories (Supervised), which is 57.7\% mIoU. However, as the number of unseen classes increases, the mIoU and hIoU gradually reduce. It is because the search space for reasoning the unseen categories is increased accordingly.

\paragraph{\textbf{Qualitative Evaluation}}
As shown in Figure \ref{fig:visual}, we qualitatively evaluate our method and the counterpart trained only on seen classes (w/o ZSS) on the ScanNet dataset, where sofa, desk, toilet and bookshelf are unseen classes, and the rest of classes are seen classes. If only training on the seen categories, the network misclassifies the unseen object into seen objects. For example, the sofa is recognized to be the chair (the first column) or table (the second column); the desk is to be the table; the toilet is to be other furniture; the bookshelf to be the wall (the fifth column) or cabinet (the last column). It shows that the network is naturally biased towards the training classes, leading to significant misclassifications for those unseen categories whose semantic representations are similar to those seen. Our method alleviates such issues and achieves decent performance in recognizing unseen objects.

\subsection{Ablation Study}
As shown in Table \ref{tab:ablation}, we conduct experiments on ScanNet 4- unseen class setting to verify the effectiveness of different components in our method, including geometric primitives-based visual representation (GP), the multi-kernels semantic representations (MK) and Unknown-aware InfoNCE Loss ($\mathcal{L}_u$). We first conduct the baseline method by following operation: 1). we remove the prototypes and directly use the point feature extracted from the backbone network as its visual representation; 2). we generate the one-kernel semantic representation from word embedding; 3). we perform visual and semantic alignment only on seen class points by the loss $\mathcal{L}_{s}$. In what follows, we present more details and insights.

\paragraph{\textbf{Effect of Geometric Primitives}}
To verify the effectiveness of the Geometric Primitives, we take the point feature extracted from the backbone network as its visual representation (Base+$\mathcal{L}_{u}$+MK$_{16}$). Compared with our full method (Base+$\mathcal{L}_{u}$+GP$_{128}$+MK$_{16}$), the unseen mIoU is dropped by about 5\%, showing the visual representation with geometric primitives beneficial for knowledge transferring from seen to unseen categories. Besides, when the geometric primitives number $M$ is 48, 96, 128 and 256 (Base+$\mathcal{L}_{u}$+GP$_{48,96,128,256}$ +MK$_{16}$), we find that too big or too small $M$ harms the performance.

\paragraph{\textbf{Effect of Semantic Representations}}
By applying the multi-kernel semantic representations (Base+$\mathcal{L}_{u}$+MK$_{16}$), the mIoU is 7\% larger than that without it (Base+$\mathcal{L}_{u}$), indicating the multi-kernel semantic representations well match the mixture-distributed visual representation. We also conduct the experiments with the different numbers of kernels, where $K$ is 2, 4, 8, 16 and 32. (Base+$\mathcal{L}_{u}$+GP$_{128}$+ MK$_{2, 4, 8, 16, 32}$). We choose $K$ to be 16 in our full method, achieving the best performance.

\paragraph{\textbf{Effect of Unknown-aware InfoNCE Loss}}
We implement multiple loss functions to constrain the unseen class points. $\mathcal{L}_{pseudo}$ indicates pseudo-labelling the unseen points via the Argmax operation on unseen categories and retrains the network with the pseudo labels. $\mathcal{L}_{self}$ is borrowed from Fixmatch \cite{sohn2020fixmatch} that imposes prediction consistency of unseen points on two augmented scenes. The last one is $\mathcal{L}_{u}$ described in formula \ref{equ:unseen_loss}. The comparison results show our Unknown-aware InfoNCE Loss loss's superiority (Base+$\mathcal{L}_{pseudo}$, Base+$\mathcal{L}_{self}$ and Base+$\mathcal{L}_{u}$).

\section{Conclusions}
We investigate transductive zero-shot semantic segmentation in this paper. Our method models the fine-grained relationship between language and geometric primitives that transfers knowledge from seen to unseen. To this end, we propose a novel visual feature representation and a novel Unknown-aware InfoNCE Loss. Therefore, under the guidance of semantic representation, the network can segment the novel objects represented by the learned geometric primitives. Extensive experiments conducted on four datasets show that our method significantly outperforms other state-of-the-art methods.

\bibliographystyle{ACM-Reference-Format}
\bibliography{acmart}


\end{document}